\pdfoutput=1

\documentclass[11pt]{article}

\usepackage{acl}

\usepackage{times}
\usepackage{latexsym}

\usepackage[T1]{fontenc}

\usepackage[utf8]{inputenc}

\usepackage{microtype}

\usepackage{tabularx} 
\usepackage{diagbox}
\usepackage{array,multirow,graphicx}
\newcommand{\STAB}[1]{\begin{tabular}{@{}c@{}}#1\end{tabular}}
\newcommand{\tabincell}[2]{\begin{tabular}{@{}#1@{}}#2\end{tabular}}

\usepackage{adjustbox}

\makeatletter
\renewcommand{\paragraph}{%
  \@startsection{paragraph}{4}%
  {\z@}{1.0mm \@plus 1ex \@minus .2ex}{-1em}%
  {\normalfont\normalsize\bfseries}%
}
\makeatother

\usepackage{arydshln}

\title{Do Prompts Solve NLP Tasks Using Natural Language?}

\author{
  Sen Yang$^{\spadesuit}$\thanks{\ \  The first two authors contributed equally to this work. }, \ \ Yunchen Zhang$^{\bigtriangleup *}$, \ \ Leyang Cui$^{\heartsuit \spadesuit}$, \ \ Yue Zhang$^{\spadesuit \Diamond}$ \\
  $^\spadesuit$School of Engineering, Westlake University \\
  $^\bigtriangleup$University of Electronic Science and Technology of China \\
  $^\heartsuit$Zhejiang University \\
  $^\Diamond$Institute of Advanced Technology, Westlake Institute for Advanced Study \\
  \texttt{senyang.stu@gmail.com} \hspace{5pt} \texttt{zhangyunchen848@gmail.com} \\
  \texttt{\{cuileyang, zhangyue\}@westlake.edu.cn}
  
  }

\begin{document}
\maketitle
\begin{abstract}
Thanks to the advanced improvement of large pre-trained language models, prompt-based fine-tuning is shown to be effective on a variety of downstream tasks.
Though many prompting methods have been investigated, it remains unknown which type of prompts are the most effective among three types of prompts (i.e., human-designed prompts, schema prompts and null prompts). 
In this work, we empirically compare the three types of prompts under both few-shot and fully-supervised settings. 
Our experimental results show that schema prompts are the most effective in general.
Besides, the performance gaps tend to diminish when the scale of training data grows large. 
\end{abstract}

\section{Introduction}

Prompt-based fine-tuning has gained increasing attention on NLP \cite{shin-etal-2020-autoprompt,schick-etal-2020-automatically,schick-schutze-2021-just,tanl,gao2021making}.
The main idea is to leverage knowledge in pre-trained language models for downstream tasks, by reformulating a specific task into the form of language modeling tasks, with the aid of prompts.
Among various recent methods on prompt-based NLP, there has been three major forms of prompts, which we call {\it NL template prompts}, {\it schema prompts} and {\it null prompts}, respectively.
NL template prompts \cite{petroni-etal-2019-language,jiang-etal-2020-know,gao2021making} were the earliest proposed and the dominant method.
As illustrated in Table \ref{tab:intro_example}, they use a natural language sentence to augment a given input, where the added prompt contains a mask token that indicates the output class.
In contrast, schema prompts \cite{lee2021dialogue,tanl} replace a natural language sentence with a structured schema, which makes the prompt more succinct and code-like.
Null prompts \cite{logan2021cutting} are the most succinct version, directly adding a masked token to the end of the input.

While different types of prompts have been compared for specific tasks \cite{gao2021making,logan2021cutting}, there has been little work systematically comparing their effects over a large variety of tasks and training settings (i.e., few shot).
We aim to fill the gap by empirically addressing the following three research questions:

\begin{table}[t!]
    \centering
    \begin{small}
    \begin{tabular}{ll}

        \hline
        \multicolumn{1}{l}{\textbf{\textit{<Input>}}} \\
        \multicolumn{1}{l}{``{\it The movie fails to live up to the sum of its parts. }''}\\
        \hline
        {\bf Template Prompt:} \\
        {\tt [CLS]} {\it <Input>} {\it It was } {\tt [MASK]} {\tt [SEP]}. \\
        \hdashline
        {\bf Schema Prompt:} \\
        {\tt [CLS]} {\it <Input>} {\it Sentiment: } {\tt [MASK]} {\tt [SEP]}. \\
        \hdashline
        {\bf Null Prompt}~\cite{logan2021cutting}{\bf :} \\
        {\tt [CLS]} {\it <Input>} {\tt [MASK]} {\tt [SEP]}. \\
        \hline
    \end{tabular}
    \caption{An example of different types of prompts, from SST-2 dataset. }
    \label{tab:intro_example}
    
    \end{small}
\end{table}

First,  which type of prompt is generally the most effective?
Intuitively, natural language prompts better connect large pre-training and task fine-tuning by having the same language style in both phases.
However, it can increase the difficulty of representation by introducing overly long sequence extensions.
In contrast, schema and null prompts are more succinct, but less close to natural language pre-training. 

Second, are task-specific information useful to include in prompts.
Compared with NL templates and schemas, null prompts are the most succinct, and are task-agnostic in not including any task hints in the augmented sequence.
While having been shown effective for several NLI-style classification tasks under few-shot settings \cite{logan2021cutting}, it remains a question whether they are competitive in more general settings.

Third, what is the effect of automatically searching for prompt template and masked label words?
There has been a line of work automatically finding prompts, which results in seemingly unnatural augmented sequences \cite{shin-etal-2020-autoprompt,gao2021making}.
In addition, the words to use for filling the masked output slots are also flexible.
We want to learn whether these automatic selections have significant benefit compared with human definitions.

Results show that among the three types of prompts, schema prompts are the most effective in general.
However, the gap between the three types of prompts tends to diminish when the scale of training data grows sufficiently large.
Finally, both automatic templates and automatic tokens give better results compared with more understandable human prompts. 
Our code will be released at https://github.com/anonymous.
\section{Experimental Setup}

\subsection{Basic Settings}
We mainly experiment on sentence classification tasks, which have been extensively investigated in previous work~\cite{schick-schutze-2021-exploiting,gao2021making}.
We also include two structure prediction tasks (i.e., NER and relation classification) for generalization beyond sentence classification. 
For sentence classification, we follow \citet{gao2021making} to adopt {\tt RoBERTa-large}~\cite{liu2019roberta} and conduct experiments on eight sentence classification datasets
For structure prediction, we use the method and setting of \citet{cui-etal-2021-template}, which formulized NER as a text generation task. 
We experiment with CoNLL03 Dataset for NER and TACRED Dataset for relation classification. 
We adopt the same hyperparameters used in previous work. 

To get a comprehensive view, we experiment with both few-shot and rich-resource settings. 
We also report standard deviation in the few-shot experiments. 
More details about experimental settings can be found in Appendix~\ref{appendix:setting}.

\begin{table}[!ht]
    \centering
    \begin{small}
    \begin{tabular}{ccc}
        \hline
         \multirow{2}{*}{\textbf{Dataset}} &  \multicolumn{2}{c}{\it\textbf{<Pattern>}}\\
         & {\bf Template} & {\bf Schema} \\
         \hline
         \tabincell{l}{SST-2} & It was  & Sentiment: \\
         \hline
         \tabincell{l}{SST-5} & It was  & Sentiment: \\
         \hline
         \tabincell{l}{MR} & It was  & Sentiment: \\
         \hline
         \tabincell{l}{CR} & It was  & Sentiment: \\
         \hline
         \tabincell{l}{MPQA} & It was  & Opinion: \\
         \hline
         \tabincell{l}{Subj} & This is  & Opinion: \\
         \hline
         \tabincell{l}{TREC} & --  &  Question type:  \\
         \hline
         \tabincell{l}{CoLA} & This is  & Grammatical: \\
         \hline
    \end{tabular}
    \caption{
        Examples of template-based and schema-based prompts of various sentence classification tasks. The prototype prompt is formulized as ``{\tt [CLS]}{\it <Input>} {\it <Pattern>} {\tt [MASK]} {\tt [SEP]}. '', except that the template prompt of TREC is ``{\tt [CLS]} {\tt [MASK]}: {\it <Input>} {\tt [SEP]}. '' which does not require any patterns. 
    }
    \label{tab:prompt_examples}

    \end{small}
\end{table}

\begin{table}[t!]
    \centering
    \begin{small}
    \begin{tabular}{l}
        \hline
        \multicolumn{1}{c}{\bf CoNLL03} \\
        \hline
        \tabincell{l}{{\bf Templae:}\\{\it <Input>}. ${\tt [span]}$ is a {\it person} entity. } \\
        \tabincell{l}{{\bf Schema:}\\{\it <Input>}. ${\tt [span]}$: {\it person} entity. } \\
         \hline
         \hline
         \multicolumn{1}{c}{\bf TACRED} \\
         \hline
         \tabincell{l}{{\bf Template:}\\{\it <Input>}. The relation between ${\tt [span_1]}$ and ${\tt [span_2]}$ \\is {\it no\_relation}. } \\
         \tabincell{l}{{\bf Schema:}\\{\it <Input>}. [ ${\tt [span_1]}$ | ${\tt [span_2]}$ ] relation: {\it no\_relation}. } \\
         \hline
    \end{tabular}
    \caption{Example prompts for structure prediction tasks, where ${\tt [span]}$ refers to a text span in the input sentence (i.e., {\it <Input>}) and the italic parts (e.g, {\it person} and {\it no\_relation}) are the entity or relation types. For more details, please refer to \citet{cui-etal-2021-template}. }
    \label{tab:prompt_examples_ner}
    \end{small}
    \vspace{-8pt}
\end{table}

\begin{table*}[ht]
    \centering
    \begin{adjustbox}{max width=\textwidth}
    \begin{tabular}{llccccccccc}
        \hline
          & & \textbf{SST-2} & \textbf{SST-5} & \textbf{MR} & \textbf{CR} & \textbf{MPQA} & \textbf{Subj} & \textbf{TREC} & \textbf{CoLA} & \multirow{2}{*}{\bf \# Wins} \\
          & & (acc) & (acc) & (acc) & (acc) & (acc) & (acc) & (acc) & (Matt.) & \\
        \hline
         \multirow{8}{*}{\STAB{\rotatebox[origin=c]{90}{{\bf Few-shot}}}} & Fine-tuning                       & 81.4 (3.8) & 43.9 (2.0) & 76.9 (5.9) & 75.8 (3.2) & 72.0 (3.8) & 90.8 (1.8) & 88.8 (2.1) & {\bf 33.9 (14.3)} & 1 \\ 
        \cline{2-11} 
         & Template Prompt                   & 92.6 (0.6) & 47.2 (1.3) & 87.1 (1.9) & 90.7 (0.9) & 84.5 (2.3) & 91.3 (1.2) & 85.8 (2.4) & 9.2 (6.7) & 0 \\ 
         & \hspace{0.5em} w/ auto label word & 92.4 (1.0) & 43.6 (1.4) & 86.3 (2.4) & 90.2 (1.1) & 85.8 (1.7) & 91.2 (1.1) & 88.7 (3.3) & 13.9 (14.3) & 0 \\ 
         & \hspace{0.5em} w/ special token   & 91.5 (1.3) & 45.5 (1.3) & 84.7 (1.5) & 86.5 (4.7) & 73.5 (6.6) & 90.4 (2.4) & 84.6 (2.9) & 11.2 (7.9) & 0 \\ 
         \cline{2-11} 
         & Schema Prompt                     & 93.2 (0.1) & {\bf 50.2 (0.7)} & 87.3 (1.0) & 91.6 (0.6) & 85.2 (1.5) & 91.4 (0.5) & 87.8 (2.2) & 9.6 (3.0) & 1 \\ 
         & \hspace{0.5em} w/ auto label word & {\bf 93.6 (0.6)} & 47.9 (1.1) & {\bf 87.5 (1.6)} & {\bf 91.7 (0.7)} & {\bf 86.0 (0.5)} & {\bf 91.9 (0.9)} & 89.2 (2.1) & 15.0 (2.5) & 5 \\ 
         & \hspace{0.5em} w/ special token   & 92.0 (1.3) & 48.8 (3.5) & 86.4 (2.1) & 87.0 (3.7) & 68.5 (3.3) & 90.4 (0.9) & {\bf 90.6 (1.4)} & 14.6 (5.2) & 1 \\ 
         \cline{2-11} 
         & {\it Null} Prompt                 & 92.7 (0.6) & 49.0 (1.1) & 86.4 (1.3) & 89.9 (0.5) & 80.6 (1.6) & 89.8 (1.3) & 86.7 (3.2) & 9.7 (6.4) & 0 \\ 
        \hline
        \hline
        \multirow{8}{*}{\STAB{\rotatebox[origin=c]{90}{{\bf Full-size}}}} & Fine-tuning                       & 95.0 & 58.7 & 90.8 & 89.4 & 87.8 & {\bf 97.0} & 97.4 & 62.6 & 1 \\ 
        \cline{2-11} 
        & Template Prompt                   & 95.1 & 59.0 & 91.2 & 91.6 & 89.8 & 95.5 & 96.8 & 66.3 & 0 \\ 
        & \hspace{0.5em} w/ auto label word & 95.6 & 58.8 & 91.7 & 91.3 & 90.8 & 95.8 & 97.0 & {\bf 68.0} & 1 \\ 
        & \hspace{0.5em} w/ special token   & 95.6 & 58.3 & 91.6 & {\bf 92.7} & 90.3 & 96.4 & 84.8 & 65.4 & 1  \\ 
        \cline{2-11} 
        & Schema Prompt                     & 95.2 & {\bf 59.4} & 91.4 & 91.8 & 90.2 & 95.6 & 97.2 & 67.0 & 1  \\ 
        & \hspace{0.5em} w/ auto label word & 95.1 & 58.4 & 91.9 & 91.9 & 89.8 & 96.7 & {\bf 97.6} & 66.4 & 1  \\ 
        & \hspace{0.5em} w/ special token   & {\bf 95.8} & 57.7 & {\bf 92.4} & 91.4 & {\bf 90.9} & 96.4 & 97.2 & 67.8 & 3 \\ 
        \cline{2-11} 
        & {\it Null} Prompt                 & 95.7 & 55.9 & 90.5 & 87.5 & 90.8 & 96.0 & 96.4 & {\bf 68.0} & 1  \\ 
        \hline
    \end{tabular}
    \end{adjustbox}
    \caption{Experiment results on sentence classification. 
            We report standard deviation for few-shot experiments. 
            The results with {\it null} prompt~\cite{logan2021cutting} are produced by our re-implementation. 
            The results with {\it null} prompts should be compared with template- and schema-based prompts without auto label word or special token. }
    \label{tab:results}
    \vspace{-8pt}
\end{table*}

\begin{table}[!ht]
    \centering
    \resizebox{0.44\textwidth}{!}{
    \begin{tabular}{lcccc}
        \hline
        & \multicolumn{2}{c}{\textbf{CoNLL03}} & \multicolumn{2}{c}{\textbf{TACRED}} \\
        & \multicolumn{2}{c}{(F1)} & \multicolumn{2}{c}{(F1)} \\
        & Few & Full & Few & Full \\
        \hline
        Fine-tuning & 34.3 (2.9) & 90.8 & 20.8 (1.1) & 64.9 \\
        \hline
        Template & 62.1 (5.1) & 90.3 & 27.2 (1.6) & 69.77 \\
        Schema   & 70.0 (2.9) & 90.7 & 28.1 (1.6) & 69.82 \\
        \hline
    \end{tabular}
    }
    \caption{Experiment results for named entity recognition and relation classification. }
    \label{tab:results_structure}
\end{table}

\subsection{Prompt}

We experiment with template prompts, schema prompts and {\it null} prompts~\cite{logan2021cutting}. 
Following ~\citet{schick-schutze-2021-exploiting}, a prompt method generally contains a {\it pattern} that maps inputs to prompt-style outputs and a {\it verbalizer} that maps labels to vocabulary tokens. 
In this paper, the term ``prompt'' normally refers to the {\it pattern}, while the term ``label word'' refers to the verbalizer.
Some prompt examples are shown in Tables \ref{tab:prompt_examples} and \ref{tab:prompt_examples_ner}.

\subsection{Prompt Types}
\label{appendix:prompt_type}
Tables~\ref{tab:prompt_examples} and \ref{tab:prompt_examples_ner} give examples for different prompt patterns. 

\paragraph{Template Prompt}
We define template prompts are fluent sentences that contains task-specific hints.
Our template-based prompts for sentence classification and structure prediction are respectively from ~\citet{gao2021making} and ~\citet{cui-etal-2021-template}. 

\paragraph{Schema Prompt}
We define schema-based prompts as syntactically-incorrect but task-related prompts.
We design schema-based prompts that are unnatural to human speaking and writing, using our intuition about the specific tasks. 
We do not further tune any of these prompts in our experiments.
Although this may introduce subjective bias to our experiments, we argue that another different set of schema-based prompts would not make significant difference when comparing with template-based prompts. 

\paragraph{Null Prompt}
\citet{logan2021cutting} proposed to use {\it null} prompts in few-shot prompt fine-tuning: the {\it pattern} is entirely removed, and only the label word is utilized.

\subsubsection{Label Word}
For sentence classification, different types of label words are investigated, including manually-designed ones, automatically searched ones~\cite{gao2021making} and special tokens.
The former two are pretrained words that are already in the vocabulary, while special tokens are those label tokens that are randomly initialized from scratch, with one token (e.g., {\tt [T0]} and {\tt [T1]}) for each label. 
We adopt the automatic label word searching method of \citet{gao2021making}.
Though other work has also investigated automatic label word generation~\cite{schick-etal-2020-automatically}, we do not include them all because the label word searching methodology is not our focus. 
\section{Analysis}



\subsection{Do Prompts Have to Follow Human Speaking  And  Writing?}
\label{sec:analysis_human}
The results for sentence classification are shown in Table~\ref{tab:results}. 
Under the few-shot setting, schema-based prompts achieve the best performance on all datasets except for CoLA.
Taking different types of prompts into comparison, schemas consistently outperform templates by $0.6\sim 2.2$ points and outperform {\it null} prompts by $0.5\sim 4.6$ points (except for CoLA), indicating that schemas are better few-shot learners. 
Under the full-size setting, although the performance gaps, ranging from $0.1$ to $0.7$ points, are not as significant as those of the few-shot setting, schema-based prompts still gain 5 wins out of 8 datasets. 

Table~\ref{tab:results_structure} gives the results of structure prediction tasks.
Under the few-shot setting, schemas give the best performance, while prompting methods are generally better than fine-tuning. 
The improvements are much more significant than those of the sentence classification tasks, with absolute improvements of 35.7 F1 for CoNLL03 and 7.3 F1 for TACRED. 
Under the rich-resource setting, fine-tuning and prompting methods are shown to be competitive with each other.

Interestingly, we observe that the standard deviations of schema-based prompts are consistently the lowest among all methods, especially for comparing with template-based ones. 
Since standard deviation mainly results from the differences among randomly sampled train and development sets across 5 runs, the low standard deviation suggests that schema-based prompts are more stable than templates when data resources are limited. 
This also suggests that the performance lower-bound of schema-based prompts is relatively high,  regardless of the quality of randomly sampled datasets. 
This advantage of schema-based prompts may be of great use in real-life scenarios, in which datasets often have very limited sizes and much noise. 

As shown in Table~\ref{tab:results}, prompting methods with automatically-searched label words achieve the best results on 5 out of 8 datasets under the few-shot setting, showing their advantage against human-designed label words. 
In particular, all these 5 wins are obtained by schema-based prompts.
Because the automatic label words are generated when the language model is not tuned yet, we conclude that pretrained LMs can already make use of grammatically-incorrect schemas even before any tuning. 
This implies that schemas might also be effective under zero-shot settings, which we leave for future investigation. 
 


\begin{table}[!t]
    \centering
    \begin{adjustbox}{max width=0.48\textwidth}
    \begin{tabular}{l|c|ccc}
        \hline
        \multirow{3}{*}{\bf Dataset} & \multirow{3}{*}{\bf Labels} & \multicolumn{3}{c}{\bf Label Words}\\
         & & {\bf Manual} & \multicolumn{2}{c}{\bf Automatic} \\
         & & Template \& Schema & Template & Schema \\
         \hline
         SST-2 & positive / negative  & great / terrible & exquisite / disgusting  & \fbox{pure / dead} \\
         \hdashline
         MR    & positive / negative & great / terrible & magical / laughable  & \fbox{brilliant / blah} \\
         \hdashline
         Subj  & subjective / objective & subjective / objective & obvious / murder &  \fbox{Nil / unknown} \\
        \hline
    \end{tabular}
    \end{adjustbox}
    \caption{Labels, manually-designed label words and automatically-searched label words for three datasets. We use the same set of manually-designed label words for both template and schema-based prompts. Those label words that obtain the best performance under the few-shot setting are \fbox{framed out}. }
    \label{tab:partial_label_word}
\end{table}

\subsection{Are Task-Specific Hints Needed?}
\label{sec:task-specific}

\paragraph{Few-shot Setting}
Though \citet{logan2021cutting} argued that {\it null} prompts could achieve competitive results compared with manually designed and automatically-generated prompts under the few-shot setting, our experiments give different results.
Taking schema-based and {\it null} prompts into comparison\footnote{For fair comparison, we compare the ``Schema Prompt'' row (without automatic label words or special tokens) and the ``{\it Null} Prompt'' row in Table~\ref{tab:results}. }, the former outperforms the latter for most few-shot tasks, with the only exception being CoLA on which the performance gap is just 0.1\%. 
Schema-based and {\it null} prompts are mainly different in that the former is augmented with task-specific hints while the latter is not.
Taking question classification task (TREC Dataset) as an example, the corresponding schema-based prompt is ``{\it <Input>} Question type: {\tt [MASK]}. '' while the {\it null} prompt is ``{\it <Input>} {\tt [MASK]}. ''. 
An absolute improvement of 1.1 \% is obtained by merely augmenting the prompt with two task-related words (i.e., ``Question type'').
Therefore, we can conclude that task-specific hints are still needed for prompt-based few-shot learning. 


\paragraph{Rich-resource Setting}
Under the fully-supervised setting, schemas and {\it null} prompts are competitive with each other and either wins for 4 out of 8 datasets. 
The impacts of task-specific hints are overridden by large amounts of training data. 
Fine-tuning methods give an even more extreme condition: the hints and the label words are both removed, but fine-tuning still shows competitive performance with prompting. 
This is in line with previous work, which suggested that prompting methods are mostly effective for zero- and few-shot settings.


\begin{table}[t!]
    \centering
    \begin{small}
    \begin{tabular}{lc}
        \hline
        \multicolumn{1}{l}{{\bf SST-2} (Label Words: great / terrible)} & {\bf Acc}  \\
        \hline
        \tabincell{l}{{\bf Manual:}} &  \\
        {\it <Input>} It was {\tt [MASK]}. & 92.6 (0.6) \\
         \hline
         \tabincell{l}{{\bf Auto Template:}} &  \\
        {\it <Input>} It's {\tt [MASK]}! & 92.7 (0.9) \\
        {\it <Input>} That's {\tt [MASK]}. & 92.6 (0.7) \\
        {\it <Input>} Its {\tt [MASK]}. & 92.4 (0.8) \\
        It's {\tt [MASK]}. {\it <Input>} & 92.1 (1.1) \\
        Absolutely {\tt [MASK]}. {\it <Input>} & 91.4 (1.4) \\
        Just {\tt [MASK]}. {\it <Input>} & 89.9 (1.6) \\
         \hline
    \end{tabular}
    \caption{Examples for automatically-searched templates and their performance on SST-2 dataset. }
    \label{tab:auto_template}
    \end{small}
    \vspace{-10pt}
\end{table}

\subsection{Automatic Search versus Manual}
\label{sec:auto}

By examining all automatically searched label words, we find that most of them that obtain superior performance tend to be unnatural (e.g., ``dead'', ``blah'' and ``Nil'' as shown in Table~\ref{tab:partial_label_word}\footnote{The full list of label words are shown in Appendix~\ref{appendix:table_label_words}}), which further verifies that prompts do not need to strictly follow the way in which humans speak and write. 
However, the automatic label words are not totally nonsense.
For example, ``dead'' and ``terrible'' both tend to be negative, and ``unknown'' and ``objective'' both mean that something is out of one's mind.
This ``loosely-connected synonym'' situation results in an assumption that effective label words should be consistent with human intuitions, though they are not strictly required to be natural or grammatically-correct.

We take SST-2 as an example to examine the effect of automatically-searched prompt templates.
As shown in Table~\ref{tab:auto_template}, the automatically generated template outperforms the manual one by adding a ``!'' at the end of the sentence, which suggests that automatic search is the optimal. 
Existing work that investigated automatic prompt search also pointed out that machine-generated prompts are superior to human-designed ones~\cite{shin-etal-2020-autoprompt,jiang-etal-2020-know,gao2021making}.

\bibliography{custom}
\bibliographystyle{acl_natbib}

\appendix

\section{Appendix}

\begin{table*}[!ht]
    \centering
    \begin{adjustbox}{max width=1.0\textwidth}
    \begin{tabular}{l|c|ccc}
        \hline
        \multirow{3}{*}{\bf Dataset} & \multirow{3}{*}{\bf Labels} & \multicolumn{3}{c}{\bf Label Words}\\
         & & {\bf Manual} & \multicolumn{2}{c}{\bf Automatic} \\
         & & Template \& Schema & Template & Schema \\
         \hline
         SST-2 & positive / negative  & great / terrible & exquisite / disgusting  & \fbox{pure / dead} \\
         \hdashline
         SST-5 & \tabincell{c}{v.pos. / positive / neutral\\ / negative / v.neg.} &  \fbox{\tabincell{c}{great / good / okay \\ / bad / terrible}} & \tabincell{c}{excellent / good / hilarious\\ / terrible / awful} & \tabincell{c}{pure / appropriate / ok\\ / low / dead} \\
         \hdashline
         MR    & positive / negative & great / terrible & magical / laughable  & \fbox{brilliant / blah} \\
         \hdashline
         CR    & positive / negative & great / terrible & astounding / worse  &  \fbox{winning / boring} \\
         \hdashline
         MPQA  & positive / negative & great / terrible & awesome / awful  &  \fbox{Good / FALSE} \\
         \hdashline
         Subj  & subjective / objective & subjective / objective & obvious / murder &  \fbox{Nil / unknown} \\
         \hdashline
         TREC  & \tabincell{c}{description / entity / abbreviation \\ / human / location / number} & \tabincell{c}{Description / Entity / Expression \\ / Human / Location / Number} & \tabincell{c}{Discussion / Scene / Response\\ / Fact / Results / Problem} & \tabincell{c}{Background / Static / Communication\\ / Criminal / Location / Numbers} \\
         \hdashline
         CoLA  & grammatical / not\_grammatical & correct / incorrect  &  fiction / now  & c / N \\
        \hline
    \end{tabular}
    \end{adjustbox}
    \caption{Labels, manually-designed label words and automatically-searched label words for each dataset. We use the same set of manually-designed label words for both template and schema-based prompts. Those label words that obtain the best performance under the few-shot setting are \fbox{framed out}. }
    \label{tab:label_word}
\end{table*}

\subsection{Detailed Experimental Settings}
\label{appendix:setting}
For sentence classification, our experiments are conducted on 8 datasets, including SST-2~\cite{sst}, SST-5, MR, CR, MPQA, Subj~\cite{subj}, TREC~\cite{trec} and CoLA~\cite{cola}. 
Similar to previous work, we adopt a masked language model to predict the label word and then adopt a verbalizer to map the label words to classification labels. 

For structured prediction tasks, we adopt a pretrained BART~\cite{bart} as \citet{cui-etal-2021-template} did.
As for the NER task, all possible text spans are enumerated as potential entity spans. 
The prompt sentences are generated by BART and the entity label is determined by comparing the summations of the log-likelihood of the generation process. 

For the few-shot experiments, we choose $K=16$, where $K$ refers to the number of examples per class for the training and development sets. 
We conduct few-shot experiments across 5 runs, using different randomly sampled train and development sets. 
We randomly sample 2,000 examples as the test set for sentence classification and use the full test set for structure prediction.

\subsection{Automatically Searched Label Words}
\label{appendix:table_label_words}

The full list of automatically searched label words are shown in Table~\ref{tab:label_word}.

\end{document}